\documentclass[conference]{IEEEtran}
\IEEEoverridecommandlockouts
\usepackage{cite}
\usepackage{amsmath,amssymb,amsfonts}
\usepackage{algorithmic}
\usepackage{graphicx}
\usepackage{textcomp}
\usepackage{xcolor}
\def\BibTeX{{\rm B\kern-.05em{\sc i\kern-.025em b}\kern-.08em
    T\kern-.1667em\lower.7ex\hbox{E}\kern-.125emX}}
\begin{document}

\title{AquaSight: Automatic Water Impurity Detection Utilizing Convolutional Neural Networks\\}

\author{\IEEEauthorblockN{Ankit Gupta}
\IEEEauthorblockA{\textit{Department of Computer Science} \\
\textit{TJHSST}\\
Alexandria, USA \\
2020agupta1@tjhsst.edu}
\and
\IEEEauthorblockN{Elliott Ruebush}
\IEEEauthorblockA{\textit{Department of Computer Science} \\
\textit{University of Maryland}\\
Bethesda, USA \\
eruebush@umd.edu}}

\maketitle

\begin{abstract}
According to the United Nations World Water Assessment Programme, every day, 2 million tons of sewage and industrial and agricultural waste are discharged into the world’s water. In order to address this pervasive issue of increasing water pollution, while ensuring that the global population has an efficient, accurate, and low-cost method to assess whether the water they drink is contaminated, we propose AquaSight, a novel mobile application that utilizes deep learning methods, specifically Convolutional Neural Networks, for automated water impurity detection. After comprehensive training with a dataset of 105 images representing varying magnitudes of contamination, the deep learning algorithm achieved a 96 percent accuracy and loss of 0.108.
Furthermore, the machine learning model uses efficient analysis of the turbidity and transparency levels of water to estimate a particular sample of water's level of contamination. When deployed, the AquaSight system will provide an efficient way for individuals to secure an estimation of water quality, alerting local and national government to take action and potentially saving millions of lives worldwide.
\end{abstract}


\section{Introduction}
Water is one of the basic necessities for life. However, pollution and environmental contaminants negatively impact water quality and make it potentially unsafe to drink for over 2 billion of the global population [10]. Pollution leads to nearly 9 million premature deaths a year, 16 percent of all deaths worldwide [4].  People in environments without easy access to purified water would benefit from technology allowing them to determine to what extent potential drinking water appears contaminated. In order to provide this technology and quickly determine the level of contamination of water based on an image, we propose AquaSight, a deep learning approach to water quality analysis that utilizes the power of modern hardware and machine learning techniques.

In machine learning, neural networks are a model whose design attempts to mirror aspects of the human brain and its biological neural networks. They form the basis of most methods of deep learning, a subset of machine learning that incorporates multiple sequential layers that data is run through in order to perform classification or pattern analysis. Deep learning has helped make progress in numerous fields such as computer vision, bioinformatics, and natural language processing.

Convolutional neural networks (CNNs) represent another subset of neural networks and deep learning. CNNs are complex organizations of nodes known as neurons that form connections as they are trained on data. CNNs perform supervised learning, as they require an individual to define specific features for a model to analyze. Thus, the creator of the model builds out the structure of the model, instantiating various layers that perform different functions and contribute to the model’s performance in different ways.

CNNs have grown in popularity in recent years as researchers and data scientists have realized their aptitude for various tasks, most importantly image classification.  In a famous example of CNN usage, Krizhevsky et al implemented an advanced CNN that achieved record accuracies on the 1.2 million image ImageNet dataset [3]. Furthermore, Ciresan et al have examined neural networks and determined that they may have even more versatility and potential for applications we have yet to discover [2]. Ciresan et al also helped pioneered the usage of GPUs for more efficient CNN training, which benefited our research as we used a GPU to accelerate training.
AquaSight is a novel system for image-based estimation of water quality. While in depth measures of water quality such as Water Quality Index (WQI) exist, AquaSight attempts to provide a quick estimate of water quality that does not require extensive chemical testing and can be performed by anyone, anywhere, and at any time. AquaSight seeks to provide an automated process that produces contamination classification akin to the Secchi3000 Secchi Disk Depth and turbidity-based water quality analysis that Toivanen et al introduced in 2013 [9].

Previous research, such as that of Schwartz and Levin in 1999, has shown a link between turbidity levels and gastrointestinal illness [8]. Consequently, we believe that analyzing turbidity of images of water represents a viable method for determining the risk that consuming that water poses. We seek to create a holistic machine learning model that simply requires an image of water in order to determine whether or not contaminants are present, improving the ease with which one can perform basic water quality analysis and determine the safety of drinking water.

AquaSight is an ongoing project. We currently are working on expanding our model to offer more precise distinctions between different types of contaminants, and we are almost done with an Android mobile application that will make our model publicly available and will interface directly with our model in order to streamline the process of testing on various water sources. In the near future, we also plan to add a component that allows the pictures collected by AquaSight to be sent to a storage database, where each week the CNN model will be retrained to ensure that our model stays up to date and as accurate as possible. We also have future plans to expand our research to analysis of microscopic water imagery to determine if potentially dangerous bacteria is present.

\section{Methods}

\subsection{Developing the Dataset}
The dataset we used was composed of 105 water images. It was collected in two parts. The first 91 images included clean water and all variations of contaminated water. The additional 14 images from the second round of data collection were all of clean water and introduced a wider range of light levels to the dataset.35 images from the original 91 were images of clean water, with 7 images of clean water colored with Light Blue, Green, Blue, Yellow, Brown, Orange, and Red. The remaining 56 images included 28 non-colored water images with contaminants added in and 28 colored water images with contaminants added in. The contaminants added, in sequence, were:
\begin{enumerate}
    \item{Sand}
    \item{Sand and Salt}
    \item{Sand, Salt, Black Pepper}
    \item{Sand, Salt, Black Pepper, Oil-Paint}
\end{enumerate}
14 of the 28 clean water images from the original data set were taken with the light off. 
The 14 additional images were all clean water and were taken at varying levels of brightness. 
Examples of images, in increasing contamination, are shown.
\linebreak
\linebreak
\includegraphics[width=8cm,height=4cm]{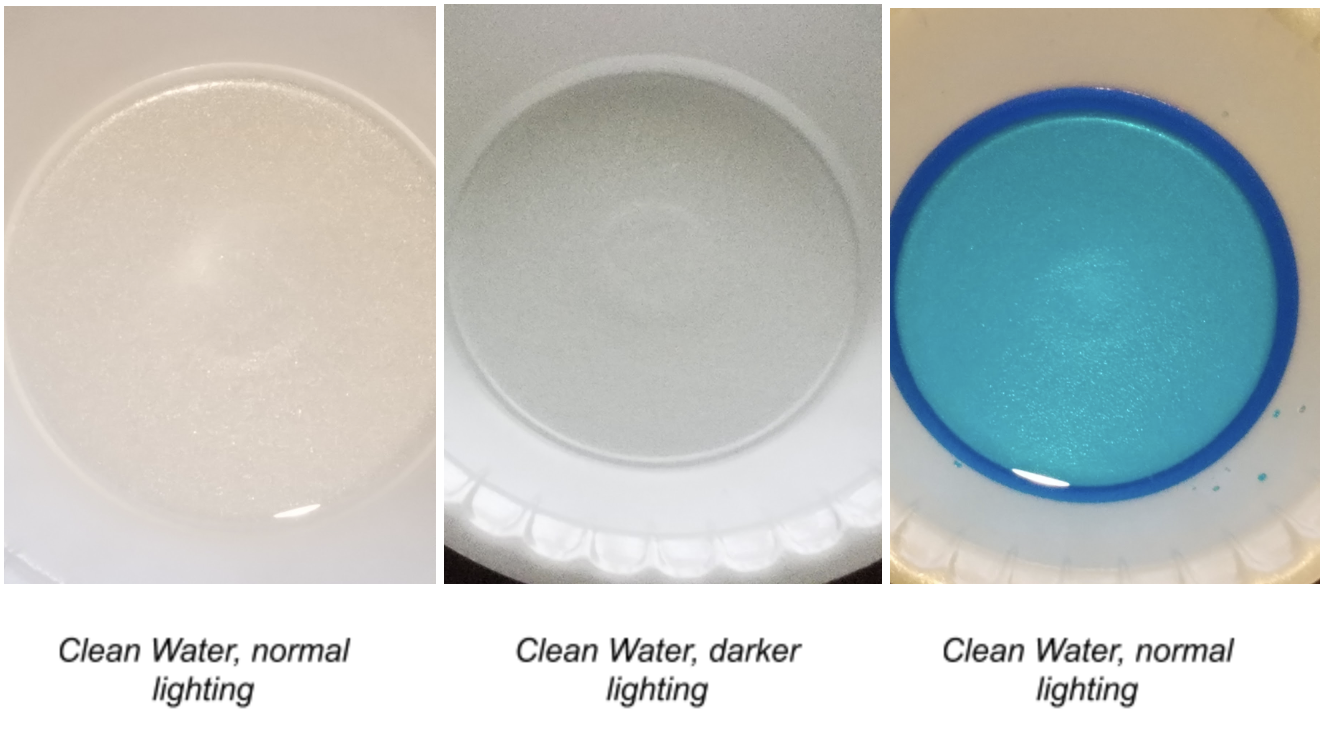}
\includegraphics[width=8cm,height=4cm]{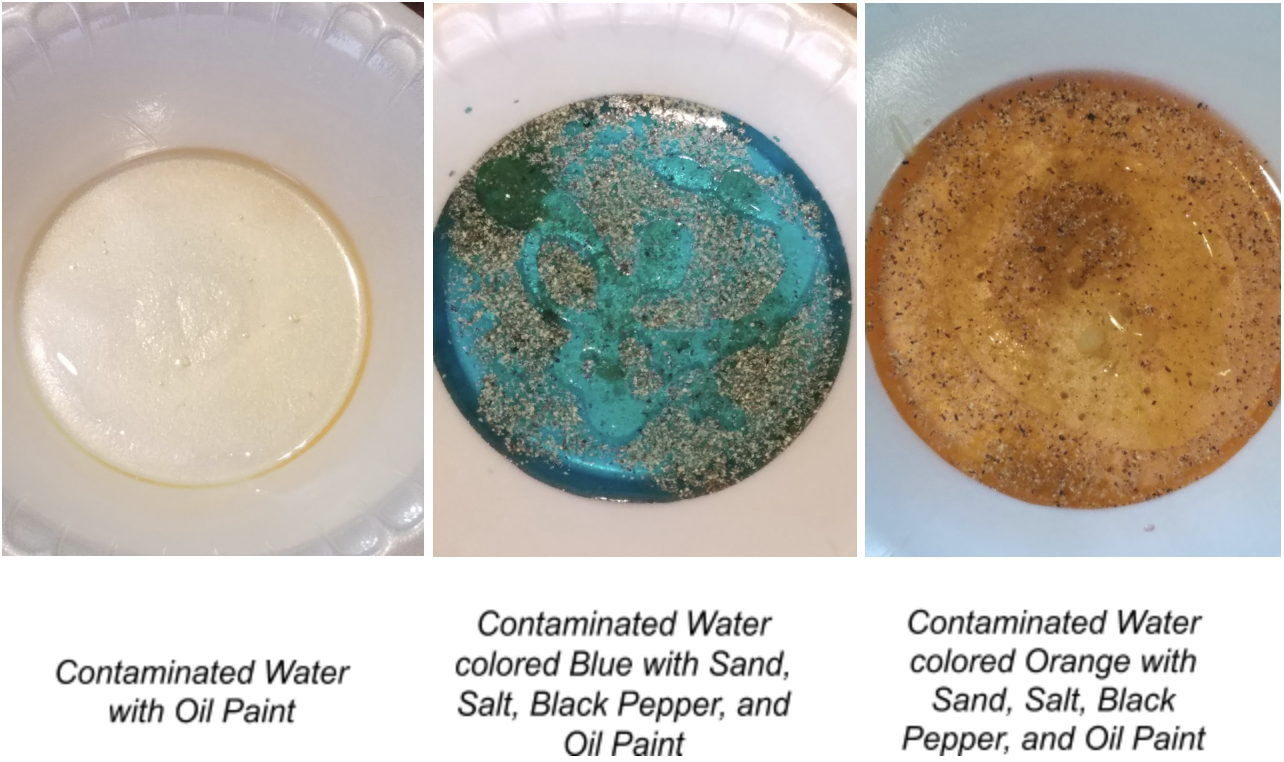}

\subsection{Training the Deep Learning Model}
Out of the 105 image dataset, 75 percent (78 images) of the dataset was used for training and 25 percent (27 images) was used for validation. In order to create our model, we used the Keras machine learning library on top of a Tensorflow backend. Using Keras, we implemented a CNN to perform a binary classification of a water image as either clean (0) or contaminated (1). As Krizhevsky et al did, we implemented dropout in our model to help avoid overfitting our model to our data. Avoiding overfitting is an especially important challenge given the relatively small size of our data set.

The network architecture is detailed below in a Keras model summary and a visualization of the model created using Netron, a deep learning and machine learning model visualizer. Predictions for each image were made on a scale of 0 to 1. If the value of the prediction rounded to 0 (val less than 0.5), then the final prediction for that image was clean. If the value of the prediction rounded to 1 (val = 0.5), then the final prediction for the image was contaminated. Therefore, the closer to 0, the more confident the model was in an image being clean, and the closer to 1, the more confident the model was in an image being contaminated.

The model was trained and tested locally on a PC running Windows 10. The NVIDIA Cuda Developer Toolkit was used with a GTX 1070 GPU and the Tensorflow-GPU library in order to accelerate model training.
\begin{center}
\includegraphics[width=9cm,height=10cm]{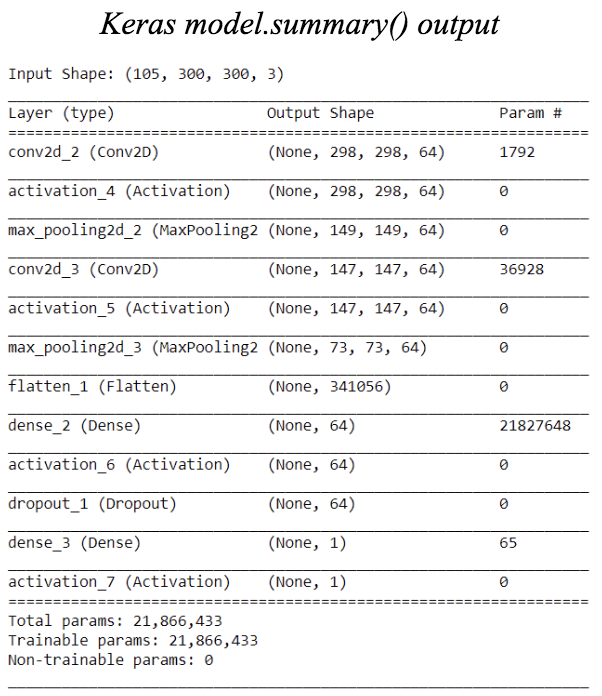}
\end{center}
In addition, below are two graphical representations of the model, include a representation created from a .h5 model file and verbose model from the .tflite model file.
\linebreak
\includegraphics[width=8cm,height=5.5cm]{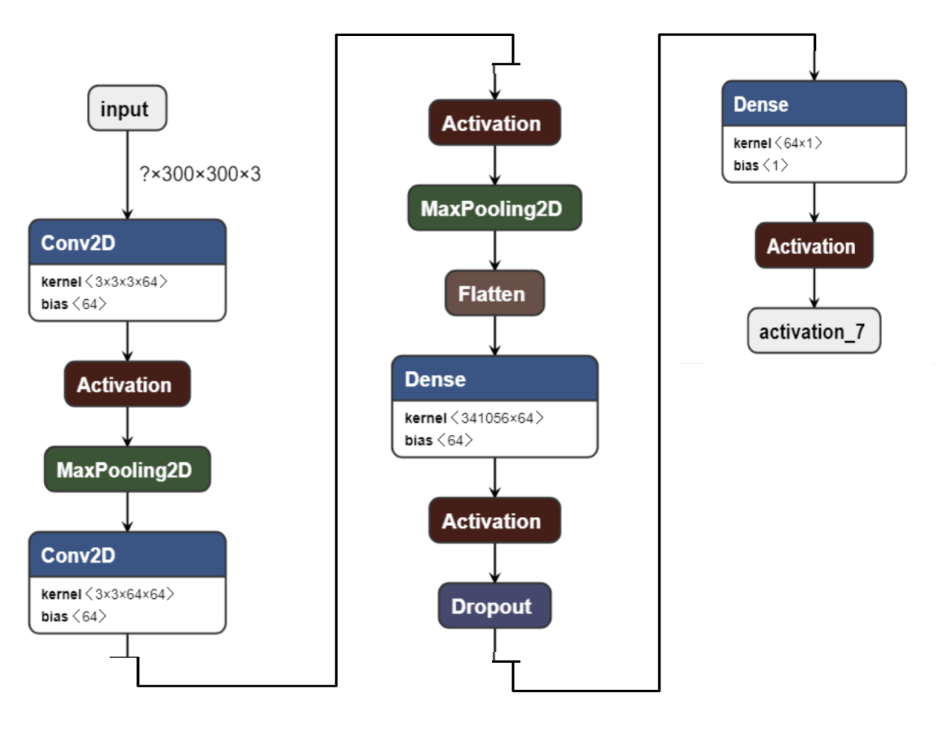}
\linebreak{}
\includegraphics[width=8cm,height=5.5cm]{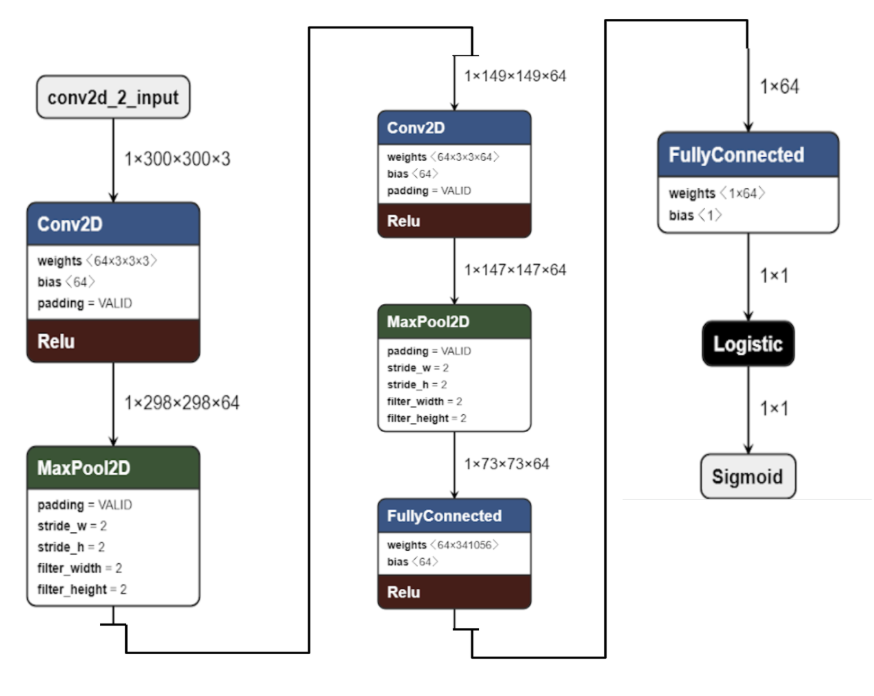}
\section{Results}
Our model had an accuracy of 96 percent (101/105 images) and a loss of .1077 after evaluating it on our image dataset. 
\linebreak{}
The resized versions of the four missed images are pictured below. Missed Image 2 represents the least confident prediction of contamination (closest to 0.5) that the model output. Three of the missed images were clean images that were incorrectly classified as contaminated. One of the missed images was a contaminated image that was incorrectly classified as clean.

The average value of clean predictions (average of all predictions for which the model output a value less than 0.5) was 0.081. The average value of contaminated predictions (average of all predictions for which the model output a value greater than 0.5) was 0.935. Predictions classified as clean ranged from 0.002 (most certain to classify as clean) to 0.463 (least certain to classify as clean). Predictions classified as contaminated ranged from 0.710 (least certain to classify as contaminated) to 0.999 (most certain to classify as contaminated).

In addition to metrics directly from the model and its prediction values, we also calculated F-beta score, Precision, Sensitivity, and Accuracy. In the formulas for these numbers, TP represents the number of true positives (correctly identifying a contaminated image as contaminated). TN is the number of true negatives (correctly identifying a clean image as clean). FP is the number of false positives (incorrectly identifying a clean image as contaminated). FN is the number of false negatives (incorrectly identifying a contaminated image as clean).

Below is a confusion matrix showing TP, TN, FP, and FN.
\linebreak{}
\begin{center}
\includegraphics[width=6cm,height=5.3cm]{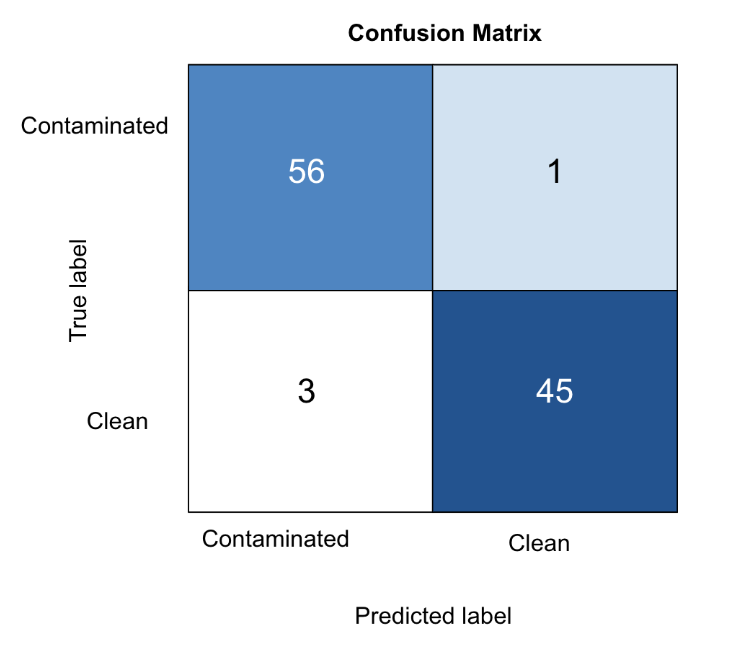}
\end{center}

There are four machine learning specific statistical measures that we used. These tests were Accuracy, Precision, Sensitivity, and Fβ (F-Beta). The equations for calculating each of the statistical measures are shown below.
\begin{center}
\includegraphics[width=4.5cm,height=3.75cm]{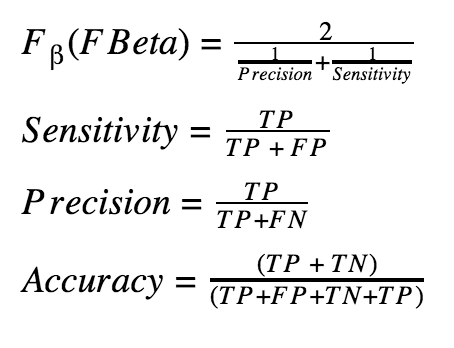}
\end{center}

The chart below shows the results of calculations for each of the four statistical measures.
\begin{center}
\begin{tabular}{ |c|c|c|c|c| } 
\hline
F-Beta & Sensitivity & Precision & Accuracy \\
\hline
0.966 & 0.949 & 0.982 & 0.962 \\ 
\hline
\end{tabular}
\end{center}

\section{Discussion}
The water environment problem that we studied was that of finding methods of classifying water as clean or contaminated that are easily accessible for individuals anywhere. This project attempts to provide assistance to people who lack a basic way to determine whether or not they are drinking contaminated water. Contaminated water represents a serious health threat, and oftentimes those who suffer as a result of contaminated water are not aware of the full extent of the danger it poses. In addition, water quality analysis can be inaccessible, especially in developing countries that do not have access to expensive chemical testing kits or other advanced methods of testing for contamination. However, our solution, AquaSight, empowers people in such situations with an easy way to analyze water in their communities with only a smartphone, one of the most ubiquitous modern devices.

\subsection{Prediction Analysis}
As mentioned in our results section, the average value of all predictions in which the CNN classified a model as clean was 0.081, and the average value of all predictions in which the CNN classified a model as clean was 0.935. Both 0.081 and 0.935 are relatively close to the values designating an image as clean or contaminated (0 or 1). As such, the CNN had a relatively high confidence in its predictions. This confidence is reflected by a high accuracy value of 96.2 percent. Interestingly, the CNN predicted more false positives (3 clean images classified as contaminated) than false negatives (1 contaminated image incorrectly classified as clean). The potential explanation for this phenomenon is that the varying degrees of darkness in the clean images caused the CNN to mispredict clean images with a lower brightness as contaminated since the image as a whole was darker and the water consequently appeared to have a higher turbidity. In order to rectify this issue and improve the CNN’s accuracy with potential nighttime images, we will manipulate the images with the OpenCV computer vision framework to normalize dark images and minimize their detrimental effect on accuracy.

Furthermore, the closer that a prediction was to 0.5, the less confident the CNN was in that prediction. This phenomenon reflects in the images that were missed. Missed Image 3 had a prediction value of 0.455, meaning it was only narrowly incorrectly identified. Additionally, Missed Image 2 had a prediction value of 0.710, which was the least confident that the CNN was in classifying an image as contaminated. These more uncertain values demonstrate that the CNN was less confident in the images that it miscategorized. As such, the miscategorized images could likely be rectified in the future with additional data and further optimization of the model. 
These results are crucial to the problem we are studying, because an accuracy of 96.2 percent means that people without access to expensive technology or complicated water testing kits will now have the ability to determine whether water is contaminated with significantly higher accuracy. This increased access to easy methods of testing can help individuals more effectively discern whether water is safe and can help avoid the ill effects of drinking unsanitary water. In addition, the sensitivity value of 0.949 and precision value of 0.982 further corroborate the potential for this technology to be reliable in a multitude of communities. Furthermore, as the number sample images in our dataset increases, the machine learning model continues to mature, resulting in fewer misclassifications and greater confidence.

\subsection{Impact on Economy}
While the technology behind our project, specifically developing a mobile application, computer vision image normalization techniques, and using convolutional neural networks for image classification, are not new, the integrated process that we propose combined with an accuracy of 96.2 percent after testing on a dataset of over 100 images is indeed novel. 
Previous researchers have investigated use of CNNs and other types of neural networks for image recognition, but very few have proposed using them specifically for water quality detection. Previous attempts at testing for water quality detection have also faced challenges in achieving high accuracy, with accuracy results typically between 60 percent and 90 percent [7]. In addition, no previous researchers have developed a holistic mobile application that allows for accurate, precise, and efficient water quality analysis and contacts local and national government organizations when a certain area is at risk for water contamination.

In the wider scientific and social context, our results are not only relevant to individuals in a community, but can also serve as tool for local and national governments. Individuals that use our mobile application will be able to download it from the Google Play store for free, then use any type of phone with a camera to automatically detect water that could have been potentially harmful, making such technology directly applicable to their lives. On a more general scale, local governments that receive water contamination alerts from local people can better warn other locals of potential dangers, further saving more people in their area from becoming ill. In addition, such communication can escalate to the national government, which can send efforts to purify water in areas where the mobile application registers multiple users receiving contamination efforts. In addition, the national government has the ability to monitor multiple local areas to most efficiently allocate resources, thus allowing for more streamlined environmental support.

\subsection{Future Research}
Our current research represents only the first steps in the application of machine learning to analyze images for water quality. In order to offer improved accuracy and offer a more complete analysis that moves beyond what is visible with the naked eye, we could expand our machine learning model to analyze microscope images of water and determine if harmful bacteria is present. Given more time and resources, we could develop a system that builds upon projects such as Peter Ma’s Clean Water AI [1]. We are interested in taking our research in this direction, and plan to introduce a more advanced version of our project in the future. Our end goal is to combine macroscopic analysis and microscopic analysis. We would like use our app as a hub to perform macroscopic analysis through the phone camera, while also allowing images collected via microscope to be uploaded for more advanced classification. However, such an ambitious project will require more data collection, more time, and more resources to complete.

\section{Conclusion}
Our CNN was successful in multiple aspects:
\begin{enumerate}
  \item The model achieved an accuracy of 96.2 percent and a loss of .108, analyzing each sample in only 4ms, much more efficiently than traditional measures of turbidity or transparency like SDD.
  \item The model produced easily understandable results, and the prediction values allowed us to analyze how confident the model was in its choice, providing a “spectrum of contamination levels.”
  \item Additional statistical measures such as F-beta, Precision, and Sensitivity all yielded uniformly strong results as well, reiterating the effectiveness of the model.
\end{enumerate}
In conclusion, this research represents a strong start to the ongoing project to develop an effective system for efficient and easy water quality analysis accessible via mobile application to those who most desperately need to determine the safety of their water sources. In the future, we hope to and expand our research to incorporate analysis of microscopic water images in order to test for the presence of dangerous bacteria. We will continue to refine the model and develop the application, leading to a mature system that allows for effective examination of water quality without complicated chemical testing procedures.

\section*{References}
[1] AI-Driven Test System Detects Bacteria in Water. (2018, March 22). Retrieved from https://software.intel.com/en-us/articles/ai-driven-test-system-detects-bacteria-in-water

[2] Ciresan, D. C., Meier, U., Masci, J., Gambardella, L. M., Schmidhuber, J. (2011, June). Flexible, high performance convolutional neural networks for image classification. \textit{In Twenty-Second International Joint Conference on Artificial Intelligence.}

[3] Krizhevsky, A., Sutskever, I., Hinton, G. E. (2012). Imagenet classification with deep convolutional neural networks. In \textit{Advances in neural information processing systems} (pp. 1097-1105).

[4] Landrigan, P. J., Fuller, R., Acosta, N. J., Adeyi, O., Arnold, R., Baldé, A. B., ... Chiles, T. (2018). The Lancet Commission on pollution and health. The Lancet, 391(10119), 462-512.

[5] Mateo-Sagasta, J., Zadeh, S. M., Turral, H. (Eds.). (2018). \textit{More people, more food, worse water?: a global review of water pollution from agriculture.} Rome, Italy: FAO Colombo, Sri Lanka: International Water Management Institute (IWMI). CGIAR Research Program on Water, Land and Ecosystems (WLE)..

[6] United Nations, Department of Economic and Social Affairs, Population Division (2017). World Population Prospects: The 2017 Revision, Key Findings and Advance Tables. Working Paper No. ESA/P/WP/248.

[7] Samantaray, A., Yang, B., Dietz, J. E., Min, B. C. (2018). Algae Detection Using Computer Vision and Deep Learning. \textit{arXiv preprint arXiv:1811.10847}.

[8] Schwartz, J., Levin, R. (1999). Drinking water turbidity and health. \textit{Epidemiology}, 86-90.

[9] Toivanen, T., Koponen, S., Kotovirta, V., Molinier, M., Chengyuan, P. (2013). Water quality analysis using an inexpensive device and a mobile phone. \textit{Environmental Systems Research}, 2(1), 9.

[10] WHO/UNICEF Joint Monitoring Programme. (2017, July 12). 2.1 billion people lack safe drinking water at home, more than twice as many lack safe sanitation [Press release]. Retrieved from https://www.who.int/news-room/detail/12-07-2017-2-1-billion-people-lack-safe-drinking-water-at-home-more-than-twice-as-many-lack-safe-sanitation

[11] Codevilla, F., Gaya, J. D. O., Duarte, N., Botelho, S. (2004). Achieving turbidity robustness on underwater images local feature detection. \textit{International journal of computer vision}, 60(2), 91-110.

[12] Ge, Z., McCool, C., Sanderson, C., Corke, P. (2015, September). Modelling local deep convolutional neural network features to improve fine-grained image classification. \textit{In 2015 IEEE International Conference on Image Processing (ICIP) (pp. 4112-4116)}. IEEE.

[13] Mahapatra, S. S., Nanda, S. K., Panigrahy, B. K. (2011). A Cascaded Fuzzy Inference System for Indian river water quality prediction. \textit{Advances in Engineering Software}, 42(10), 787-796.

[14] Yuan, F., Huang, Y., Chen, X., Cheng, E. (2018). A Biological Sensor System Using Computer Vision for Water Quality Monitoring. \textit{IEEE Access}, 6, 61535-61546.

[15] Zhang, Y., Pulliainen, J., Koponen, S., Hallikainen, M. (2002). Application of an empirical neural network to surface water quality estimation in the Gulf of Finland using combined optical data and microwave data. \textit{Remote sensing of environment}, 81(2-3), 327-336.

\end{document}